\documentclass[letterpaper, 10 pt, conference]{ieeeconf}  

\IEEEoverridecommandlockouts                              

\usepackage{amsmath} 

\usepackage{graphicx}
\usepackage{tabularx}
\usepackage{dblfloatfix}
\usepackage{placeins}
\usepackage{pifont}
\usepackage{multirow}
\usepackage{url}
\usepackage{xurl}
\usepackage[inkscapelatex=false]{svg}
\usepackage{subcaption}
\usepackage{booktabs}
\usepackage{colortbl}

\title{\LARGE \bf
Is Success All You Need? Investigating the Impact of Input Perturbations on VLA Behaviour in Tabletop Manipulation Tasks}

\author{Sophie Higham$^{1}$, Riccardo Andrea Izzo$^{2}$, Matteo Matteucci$^{2}$ and Alessandro Suglia$^{1}$ %
\thanks{$^{1}$S. Higham and A. Suglia are with the School of Informatics,
        University of Edinburgh, UK
        {\tt\small \{s.higham, asuglia\}@ed.ac.uk}. The work of S. Higham is supported by a studentship from the UKRI AI Centre for Doctoral Training in Dependable and Deployable AI for Robotics (CDT-D2AIR), grant number EP/Y030834/1. $^{2}$R.A. Izzo and M. Matteucci are with the Department of Electronics, Informatics and Bioengineering, Politecnico di Milano, Italy {\tt\small \{riccardo.izzo, matteo.matteucci\}@polimi.it}. The work of R.A. Izzo was supported by the Italian Ministry of University and Research (MUR), funded by the European Union – NextGenerationEU (PNRR – M4C2, Inv. 3.3 – D.M. 630/2024), and co-funded by Oversonic Robotics s.r.l. Benefit Company (CUP D43C24001670008; Scholarship n. 40-033-16-DOT1316508-11141). This project was supported by the Edinburgh
    International Data Facility (EIDF) and the Data-Driven Innovation Programme at the University of Edinburgh. We would like to thank Gianluca Bardaro for his valuable feedback on the manuscript.}}

\begin{document}

\bstctlcite{BSTcontrol}

\maketitle
\thispagestyle{empty}
\pagestyle{empty}

\begin{abstract}

Vision-Language-Action (VLA) models have achieved high task success rates on robot manipulation task benchmarks. More recently, there has been an emphasis on evaluating the \emph{robustness} of VLA models to perturbations. However, this robustness is still predominantly measured through Task Success Rate (TSR). In this work, we propose a benchmark-agnostic evaluation framework to measure the \emph{behavioural} robustness of models by characterising how \emph{successful} trajectories are executed under perturbation. We implement this methodology by extending the widely-used LIBERO and LIBERO-Plus benchmarks. Across three state-of-the-art VLA models, four LIBERO task suites and seven perturbation conditions, we evaluate changes in both typical successful behaviour and its variability, including metrics of motion smoothness, efficiency and gripper behaviour. We find that perturbations can alter the behaviour of \emph{successful} trajectories, a phenomenon which cannot necessarily be inferred from TSR alone. Across LIBERO suites, we identify cases where state-of-the-art VLA models achieve comparable TSR under the same perturbation condition, yet behaviour on successful trajectories diverges substantially. Therefore, to have a more robust assessment of task performance, we argue that suitable measures of robustness should capture not only whether a task is completed, but also how the robot \emph{behaves} while completing it. When evaluating the robustness of VLA models, TSR may be complemented by \emph{behavioural} evaluation metrics that characterise the nature and variability of successful task execution by robots.
\end{abstract}

\section{INTRODUCTION}

\begin{figure}[!t]
    \centering

    \includegraphics[width=\linewidth]{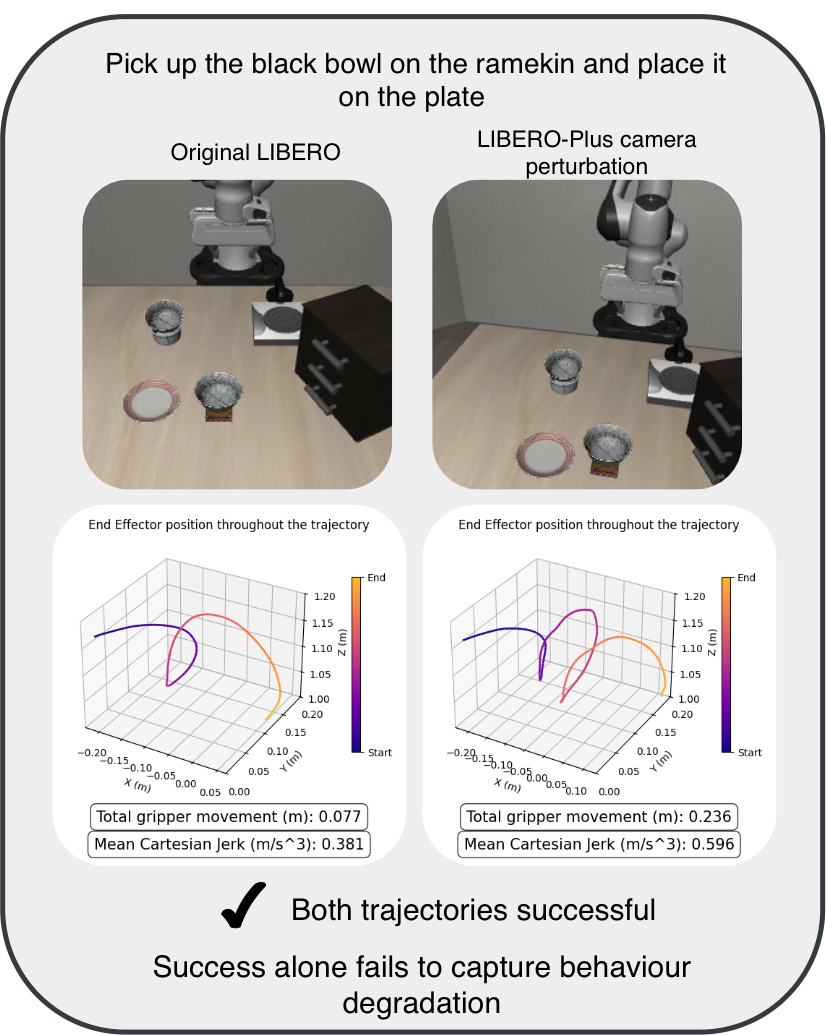}

    \caption{An illustrative example of two successful trajectories, collected using the $\pi_{0.5}$ model on the ``pick up the black bowl on the ramekin and place it on the plate" task from the LIBERO-Spatial suite. Under the LIBERO-Plus camera angle perturbation, the goal condition is met, yet the bowl is dropped and re-picked during the trajectory. TSR alone does not capture this behavioural impact; instead, it is reflected in the mean Cartesian jerk and cumulative gripper movement trajectory metrics.}
    \label{originalvsperturb}
\end{figure}

Vision-Language-Action (VLA) models \cite{brohan2023rt1roboticstransformerrealworld} have emerged as a prominent paradigm in robotics, with the ability to map visual observations, natural language instructions and proprioceptive states directly to motor commands. Building upon large-scale pre-trained Vision-Language Models (VLMs), VLAs have progressed rapidly in recent years, demonstrating generalisation to unseen objects and language instructions \cite{brohan2023rt2visionlanguageactionmodelstransfer, black2026pi0visionlanguageactionflowmodel}. Such models have shown strong performance across robotic tasks, including tabletop manipulation \cite{intelligence2025pi05visionlanguageactionmodelopenworld, wu2026vlanextrecipesbuildingstrong, kim2025finetuningvisionlanguageactionmodelsoptimizing}. More recently, increasing attention has been given to the \emph{robustness} of VLA models, which have shown brittleness to environmental variations such as altered scene lighting and visual distractors \cite{liberoplus, liberopro}. A number of works investigate this by introducing perturbations to the evaluation data, along dimensions including scene lighting and language instruction \cite{liberoplus, liberopro, morgan2026colosseumv2benchmarkinggeneralization, guo2026robustvlarobustnessvisionlanguageactionmodel}. However, this \emph{robustness} is typically measured by binary task success rate (TSR). Focussing primarily on TSR enables straightforward comparison; however, it may obscure other critical failures which aren't captured by this binary outcome alone \cite{rabanser2026scienceaiagentreliability, wang2026roboevalroboticmanipulationmeets}.

In their work on AI agent reliability, Rabanser et al. argue that once agents are deployed in the real world, they will inevitably face environmental perturbations, but that \emph{"robust agents should maintain comparable performance despite such perturbations"} \cite{rabanser2026scienceaiagentreliability}. We argue that, for \textbf{embodied AI} systems, the notion of \emph{comparable performance} should account for the fundamentally \emph{embodied} nature of the application. The decisions and control outputs of embodied AI agents translate into \emph{physical actions}, which can have significant real-world consequences \cite{Roy2021FromML}. Beyond binary TSR alone, behavioural metrics may more suitably reflect the \emph{deployability} of VLA models. We consider a number of such metrics. The \emph{jerkiness} of the trajectory may be important, as large acceleration changes can lead to higher mechanical stress and manipulator wear \cite{10610437, 7069580}. Behavioural metrics also matter from a human acceptability perspective. Sheidlower et al. \cite{10.1145/3776734.3794505} found that task completion speed data, a marker of efficiency, was requested by users. Additionally, perceived reliability and predictability are central factors affecting trust \cite{alhaji2021trust}, and unexpected behaviour from robots can degrade the trust that users may have in the technology \cite{lyons2023explanationsTrust}. In addition to TSR, we argue for the inclusion of \emph{behavioural metrics}, which we find can offer greater insight into the comparative robustness of VLA models. 

We identify a gap in the evaluation of VLA and embodied AI models. Work exists which either (i) evaluates the trajectory behaviour beyond binary task success, or (ii) measures robustness to perturbations using task success as the primary metric. To our knowledge, these two practices have not yet been combined. As a result, current robustness evaluation cannot determine whether a policy's behaviour has meaningfully degraded when it continues to complete a task successfully. \textbf{Two policies may achieve identical success rates under perturbation, while exhibiting substantially different trajectory quality}. A demonstrative example is given in Figure \ref{originalvsperturb}. Motivated by this gap, we propose a framework for evaluating the \emph{behavioural} robustness of VLA models under perturbation by calculating metrics of motion smoothness, efficiency and gripper behaviour, to complement TSR. With this evaluation framework, we aim to answer the following research questions:

\begin{itemize}

\item \textbf{RQ1:} How does the typical behaviour of \emph{successful} trajectories change under input perturbation?
\item \textbf{RQ2:} Are perturbation-induced behavioural changes reflected by changes in the TSR?
\item \textbf{RQ3:} Do perturbations alter the behavioural consistency of
\emph{successful} trajectories?

\end{itemize}

We implement this framework by extending the LIBERO \cite{libero} and LIBERO-Plus \cite{liberoplus} benchmarks. While we use these benchmarks as an experimental testbed given their wide adoption \cite{kim2025finetuningvisionlanguageactionmodelsoptimizing, liberoplus}, the proposed behavioural evaluation is benchmark-agnostic and could be applied to other benchmarks, as long as trajectory-level state and action information can be retrieved (e.g. MetaWorld) \cite{yu2021metaworldbenchmarkevaluationmultitask}. 

Thus, through this work, we:
\begin{itemize}
\item Propose a benchmark-agnostic behavioural evaluation methodology that complements task success rate with trajectory-level behaviour metrics, enabling practitioners to derive a more complete performance assessment of the robustness of VLA models, and implement it on the popular LIBERO and LIBERO-Plus benchmarks. The metrics comprise the \emph{duration}, \emph{total gripper movement}, \emph{Cartesian jerk}, \emph{joint jerk}, \emph{Cartesian path length}, and \emph{joint path length} observed for the trajectories.
\item Characterise perturbation effects in terms of the change in typical behaviour and consistency of behaviour.
\item Demonstrate that for state-of-the-art VLA models, robustness of task success does not necessarily equate to robustness of successful execution behaviour. Under perturbations, successful trajectories can change substantially even when task success is preserved.

\end{itemize}

\section{RELATED WORK}

\subsection{Vision-Language-Action models}

Building on advances in large language models (LLMs) and vision-language models (VLMs), Vision-Language-Action (VLA) models extend multimodal language modelling by predicting robot control actions from visual observations and language instructions \cite{brohan2023rt2visionlanguageactionmodelstransfer, driess2023palm}. Early VLA approaches represented robot control actions as discrete tokens alongside natural language tokens \cite{brohan2023rt2visionlanguageactionmodelstransfer}. OpenVLA is an open-source model which adopted token-based action representation, and used large-scale robot demonstration data for pre-training \cite{kim2024openvlaopensourcevisionlanguageactionmodel}. More recent approaches have used continuous action representation and action chunking. OpenVLA-OFT \cite{kim2025finetuningvisionlanguageactionmodelsoptimizing} replaces OpenVLA's discrete action representation with continuous action prediction, training using an L1 regression objective. Models in the $\pi_0$ family use flow matching for continuous action generation \cite{black2026pi0visionlanguageactionflowmodel, intelligence2025pi05visionlanguageactionmodelopenworld}. The VLANeXt open-source model similarly uses a dedicated policy module with action chunking and flow matching, conditioned on learned representations from a VLM backbone \cite{wu2026vlanextrecipesbuildingstrong}. These developments have enabled increasingly capable models, which has motivated efforts to more rigorously evaluate their behaviour and robustness. 

\subsection{Evaluation of VLA models}
\subsubsection{The focus on TSR}

We observe that the majority of widely used benchmarks for evaluating the performance of embodied AI models emphasise the overall task success rate (TSR). See Table \ref{tab:evalembodied} for a breakdown of this. Benchmarks include those in the LIBERO family (including LIBERO-Plus \cite{liberoplus} and LIBERO-PRO \cite{liberopro}). The primary reported metric is the percentage of successful episodes, measured using task-specific goal logic. Other relevant robotic manipulation benchmarks include RLBench \cite{james2019rlbenchrobotlearningbenchmark}, CALVIN \cite{mees2022calvinbenchmarklanguageconditionedpolicy}, RoboCasa \cite{nasiriany2024robocasalargescalesimulationeveryday}, MetaWorld \cite{yu2021metaworldbenchmarkevaluationmultitask} and MolmoSpaces-Bench \cite{kim2026molmospaces}. These benchmarks vary in terms of which skills are being evaluated and task complexity, however a commonality is that binary task success rates are reported. A number of works propose behaviour quality metrics beyond TSR, which can largely be grouped into metrics focussing on \emph{safety} and metrics focussing on \emph{efficiency}.  

\emph{\textbf{Safety}} In their RoboEval work, Wang et al. \cite{wang2026roboevalroboticmanipulationmeets} propose computing and reporting on \emph{joint jerk}, \emph{Cartesian jerk}, \emph{self-collisions}, \emph{environment collisions} and \emph{object slips}, arguing that evaluating TSR alone may conceal critical failures. 

\emph{\textbf{Efficiency}} Wang et al. \cite{wang2026roboevalroboticmanipulationmeets} also propose the metrics of \emph{trajectory length}, \emph{completion time}, \emph{joint path length} and \emph{Cartesian path length}. In their BEHAVIOR benchmark \cite{srivastava2022behavior}, Srivastava et al. incorporate a number of efficiency metrics, citing deployment infeasibility of policies which are overly inefficient. Specifically, they report on \emph{time taken}, \emph{kinematic displacement} (i.e., the displacement caused by an agent in the environment), \emph{logical disarrangement} (i.e., the number of changes caused by the agent to the logical state of the environment), \emph{distance navigated} (i.e., the accumulated distance travelled by the agent) and the \emph{displacement of hands} (i.e., the accumulated displacement of each hand while manipulating objects). 

\subsubsection{Introducing perturbations to measure robustness}

A number of works introduce perturbations to modify the evaluation data, and evaluate model robustness to distribution shift. Rather than evaluating models on tasks which are nearly identical to the training tasks, the LIBERO-Plus \cite{liberoplus} benchmark introduces controlled perturbations along seven dimensions. These are the \emph{object layout}, \emph{camera viewpoints}, \emph{robot initial state}, \emph{language instruction}, \emph{lighting conditions}, \emph{background textures} and \emph{sensor noise}. The authors find that, while several SOTA VLA models perform very well on the original LIBERO benchmark, under perturbations the performance of the models dropped substantially. Similarly, the LIBERO-PRO benchmark from Zhou et al. \cite{liberopro} introduces variations along four dimensions: object attributes (e.g. colour, textures), initial configuration, task instruction, and environment (e.g. background, lighting). Constraints ensure the tasks remain feasible despite the perturbations introduced, yet the authors find that models fail under simple perturbations. 

Beyond the LIBERO benchmark family, the Colosseum V2 benchmark \cite{morgan2026colosseumv2benchmarkinggeneralization} specifically targets the evaluation of VLA models, expanding and creating visual, language and action perturbation categories. In their RobustVLA work, Guo et al. \cite{guo2026robustvlarobustnessvisionlanguageactionmodel} implement four perturbation dimensions: observation, environment, language, and \emph{action}. However, these benchmarks investigate the effect of the perturbation in terms of the reduction in TSR. To the best of our knowledge, there is no work yet addressing whether \emph{successful} trajectories change as a result of the perturbations. 

\begin{table}[t]
    \centering
    \caption{Embodied AI evaluation approaches.}
    \label{tab:evalembodied}
    \small
    \setlength{\tabcolsep}{3pt}
    \renewcommand{\arraystretch}{1.1}
    \begin{tabularx}{\columnwidth}{@{}l *{3}{>{\centering\arraybackslash}X}@{}}
        \toprule
        \textbf{Work} & \textbf{TSR} & \footnotesize\bfseries\shortstack{Behavioural\\metrics} & \footnotesize\bfseries\shortstack{Robustness\\perturbations} \\
        \midrule
        LIBERO \cite{libero} & \ding{51} & \ding{55} & \ding{55} \\
        LIBERO-Plus \cite{liberoplus} & \ding{51} & \ding{55} & \ding{51} \\
        BEHAVIOR \cite{srivastava2022behavior} & \ding{51} & \ding{51} & \ding{55} \\
        RoboEval \cite{wang2026roboevalroboticmanipulationmeets} & \ding{51} & \ding{51} & \ding{55}\textsuperscript{*} \\
        RLBench \cite{james2019rlbenchrobotlearningbenchmark} & \ding{51} & \ding{55} & \ding{55} \\
        CALVIN \cite{mees2022calvinbenchmarklanguageconditionedpolicy} & \ding{51} & \ding{55} & \ding{55} \\
        RoboCasa \cite{nasiriany2024robocasalargescalesimulationeveryday} & \ding{51} & \ding{55} & \ding{55} \\
        MolmoSpaces-Bench \cite{kim2026molmospaces} & \ding{51} & \ding{55} & \ding{51} \\
        MetaWorld \cite{yu2021metaworldbenchmarkevaluationmultitask} & \ding{51} & \ding{55} & \ding{55} \\
        LIBERO-PRO \cite{liberopro} & \ding{51} & \ding{55} & \ding{51} \\
        Colosseum V2 \cite{morgan2026colosseumv2benchmarkinggeneralization} & \ding{51} & \ding{55} & \ding{51} \\
        RobustVLA \cite{guo2026robustvlarobustnessvisionlanguageactionmodel} \dag& \ding{51} & \ding{55} & \ding{51} \\
        \midrule
        \textbf{Our approach} & \ding{51} & \ding{51} & \ding{51} \\
        \bottomrule
    \end{tabularx}
    \parbox{\columnwidth}{\vspace{2pt}\footnotesize\textsuperscript{*}\,RoboEval considers spatial variations only. \dag The RobustVLA work presents a training regime for handling input/output perturbations.}
\end{table}

\section{BEHAVIOURAL ROBUSTNESS EVALUATION}

Taking inspiration from Rabanser et al. \cite{rabanser2026scienceaiagentreliability}, we consider embodied AI robustness as the extent to which \emph{successful} execution behaviour is maintained under perturbation. Deployed robotic systems will likely encounter frequent sources of environment variation and perceptual changes \cite{yoneda2022invariance}. As such, we propose a methodology which enables behavioural evaluation under input perturbations. We evaluate only successful trajectories, since the focus of our analysis is on uncovering \emph{what is masked when TSR alone is reported}. A model may be considered suitable for deployment when a high TSR is reached under perturbation conditions, but we are interested in exploring whether these ``successful" trajectories are meaningfully affected. Thus, for our evaluation purposes, we consider model behaviour under original, unperturbed tasks as the \emph{baseline}, and model behaviour under perturbed tasks as the \emph{condition}.

\subsection{Trajectory-level behaviour evaluation metrics} \label{metricsdesc}

Where applicable, we adopt metrics established by Wang et al. in their RoboEval work \cite{wang2026roboevalroboticmanipulationmeets}, including the \emph{Cartesian jerk}, \emph{joint jerk}, \emph{Cartesian path length}, \emph{joint path length}, and supplement with \emph{total gripper movement}: 

\paragraph{Duration ($s$)} The trajectory duration is calculated as elapsed simulator time between the first and final recorded control step, using the simulator control frequency. This was used, rather than wall-clock time, to isolate task completion time and prevent conflation with GPU contention or model inference latency. \footnote{Wall-clock completion time could alternatively be used, to incorporate model inference latency, potentially favouring smaller or faster models.}

\paragraph{Total gripper movement ($m$)} 
For a symmetric two-finger gripper, such as the Franka Panda, total gripper movement is the cumulative absolute change in gripper width, where width is the sum of the two finger positions. This metric captures total open/close gripper motion across the trajectory.

\paragraph{Cartesian jerk ($m/s^3$)} This metric computes the third derivative of the Cartesian position, capturing how abruptly the robot's end-effector motion changes and therefore capturing the smoothness of the trajectory \cite{wang2026roboevalroboticmanipulationmeets}. The mean Cartesian jerk can be calculated to capture average behaviour:

\begin{equation}
\label{eq:cartesian_jerk}
\resizebox{0.91\columnwidth}{!}{$\displaystyle
Jerk_{\mathrm{cart}}^{Mean}
=
\frac{1}{T-3}
\sum_{t=1}^{T-3}
\left\|
\frac{
\mathbf{x}_{t+3}
-3\mathbf{x}_{t+2}
+3\mathbf{x}_{t+1}
-\mathbf{x}_t
}
{(\Delta t)^3}
\right\|_2
$}
\end{equation}
in addition to the P95 Cartesian jerk, to capture upper-tail behaviour:

\begin{equation}
\resizebox{0.91\columnwidth}{!}{$\displaystyle
\label{eq:cartesian_jerk_p95}
J_{\mathrm{cart}}^{95}
=
Q_{0.95}
\left(
\left\|
\frac{
\mathbf{x}_{t+3}
-3\mathbf{x}_{t+2}
+3\mathbf{x}_{t+1}
-\mathbf{x}_{t}
}{
(\Delta t)^3
}
\right\|_2
\right)_{t=1}^{T-3}
$}
\end{equation}

\paragraph{Joint jerk ($rad/s^3$)}

Similarly, the joint jerk metric captures the smoothness of movement of the arm as a whole \cite{wang2026roboevalroboticmanipulationmeets}:

\begin{equation}
\label{eq:joint_jerk}
\resizebox{0.91\columnwidth}{!}{$\displaystyle
Jerk_{\mathrm{joint}}^{Mean}
=
\frac{1}{T-3}
\sum_{t=1}^{T-3}
\left\|
\frac{
\mathbf{q}_{t+3}
-3\mathbf{q}_{t+2}
+3\mathbf{q}_{t+1}
-\mathbf{q}_t
}
{(\Delta t)^3}
\right\|_2
$}
\end{equation}

\paragraph{Cartesian path length $m$}

Captures how far the end effectors of the robot arm move through physical space \cite{wang2026roboevalroboticmanipulationmeets}. 

\paragraph{Joint path length $rad$}

This is the same calculation, but in joint space, capturing how much the robot arm configuration changes over the trajectory \cite{wang2026roboevalroboticmanipulationmeets}:

\begin{equation}
\resizebox{0.91\columnwidth}{!}{$\displaystyle
\label{eq:cartesian_path_length}
L_{\mathrm{cart}}
=
\sum_{t=1}^{T-1}
\left\|
\mathbf{x}_{t+1} - \mathbf{x}_{t}
\right\|_2  \quad L_{\mathrm{joint}}
=
\sum_{t=1}^{T-1}
\left\|
\mathbf{q}_{t+1} - \mathbf{q}_{t}
\right\|_2
$}
\end{equation}

\subsection{Quantifying typical behaviour change} \label{quantifytypical}

For each manipulation task and condition, each successful rollout contributes one value for each of the behavioural metrics described in Section \ref{metricsdesc}. This creates a distribution of trajectory-level metric values for the baseline condition, and a corresponding distribution for each perturbation condition. 

For task $k$, metric $m$, and perturbation $p$, we first calculate the mean metric value across the distribution of successful trajectories $\bar{x}_{k,m,p}$. We then quantify the change in this mean value, relative to the \emph{no-perturbation} baseline $base$:

\begin{equation}
\Delta_{k,m,p}=100\frac{
\bar{x}_{k,m,p}-\bar{x}_{k,m,base}
}{
\bar{x}_{k,m,base}
}
\end{equation}

This produces one task-level percentage change in the mean metric value for each task. To summarise the effect of a perturbation condition $p$ across an entire task suite, we report the \emph{median} of these task-level percentage changes,

\begin{equation}
\widetilde{\Delta}_{m,p}
=
\operatorname{median}_{k}
\left(
\Delta_{k,m,p}
\right)
\end{equation}

This final value therefore represents the median task-level percentage change in the mean trajectory metric. Given this is an aggregation, we also report on the number of tasks in each suite for which the perturbed ($X_{k,m,p}$) and baseline ($X_{k,m,base}$) distributions differ significantly, with an increase in the mean metric value. Statistical significance is assessed using a two-sided Mann-Whitney U test \cite{mann1947test}, with Benjamini-Hochberg false discovery rate (FDR) correction \cite{benjamini1995controlling} across tasks ($\alpha=0.05$). 

\subsection{Quantifying behaviour variability} \label{quantifyvariability}

Additionally, using these distributions of trajectory-level metric values, behavioural variability can be descriptively characterised using the median absolute deviation (MAD). For each task $k$, metric $m$, and condition $p$, MAD is calculated as:

\begin{equation}
\resizebox{0.91\columnwidth}{!}{$\displaystyle
MAD_{k,m,p}=\operatorname{median}_{i}
\left(
\left|
X_{k,m,p,i} - \operatorname{median}_{j}\left(X_{k,m,p,j}\right)
\right|
\right)
$}
\end{equation}

where $X_{k,m,p,i}$ represents the metric value for each trajectory $i$ in corresponding task $k$ and condition $p$. Once a MAD value has been calculated for each task under the baseline and perturbation conditions, the percentage change in MAD for each task is calculated as:

\begin{equation}
\Delta MAD_{k,m,p} = 100\frac{MAD_{k,m,p}-MAD_{k,m,base}}{MAD_{k,m,base}}
\end{equation}

The median of these task-level percentage changes can then be reported across all the tasks in each suite. Positive values indicate increased behavioural variability, whereas negative values indicate reduced behavioural variability. 

\section{EXPERIMENTAL SETUP}

We implement this behavioural robustness framework by extending the LIBERO~\cite{libero} and LIBERO-Plus~\cite{liberoplus} benchmarks. Full code for our implementation is made available \footnote{https://github.com/esgi-research-group/vla-reliability}. Additionally, we make use of the open-source \emph{vla-eval} VLA evaluation harness from Choi et al., \cite{choi2026vlaeval}, which supports reproducible evaluation of VLA models on a variety of benchmarks and enables parallelised execution.  

We extend the trajectory data which the harness makes available. We surface the end effector position coordinates, joint positions, and gripper movement throughout each trajectory. Our post-hoc trajectory analysis pipeline processes this trajectory data and calculates per-episode values for each of the metrics introduced in Section~\ref{metricsdesc}. All trajectory derivatives are computed using the simulator control frequency of 10~Hz. These metrics are first calculated on an episode level, and then a summary for all episodes for each LIBERO task is produced. 

For the VLA models evaluated, rollouts were collected for each of the four LIBERO suites (\emph{LIBERO-Spatial}, \emph{LIBERO-Long}, \emph{LIBERO-Goal} and \emph{LIBERO-Object}) under each of the 8 perturbations (no perturbation + each of the 7 perturbation classes from LIBERO-Plus). Following standard conventions \cite{ kim2025finetuningvisionlanguageactionmodelsoptimizing, choi2026vlaeval, kim2024openvlaopensourcevisionlanguageactionmodel}, for the baseline (original) LIBERO evaluation, 50 rollouts were run for each of the 10 tasks. For each perturbation condition, we evaluate all corresponding LIBERO-Plus task variants, performing two rollouts per variant to increase the number of trajectory-level observations, with the exception of the object perturbation, for which only a single initial state is provided. 

The methodology described in Sections \ref{quantifytypical} and \ref{quantifyvariability} is applied to the LIBERO and LIBERO-Plus trajectories. Each LIBERO-Plus task variant is matched to the corresponding original LIBERO task, which serves as the baseline. Successful trajectories are identified using LIBERO's standard binary task success criterion. For each successful trajectory, the behavioural metrics described in Section \ref{metricsdesc} are calculated, and the original and perturbed distributions are compared according to Sections \ref{quantifytypical} and \ref{quantifyvariability}.

\subsection{VLA models evaluated}

Our analysis framework relies on collecting \emph{successful} trajectories, to investigate behavioural degradation which TSR alone may be masking. Thus, we targeted VLA models which had high enough TSR that there were enough samples to enable a robust comparison. A number of VLAs, which perform well on LIBERO but drastically degrade under one or more LIBERO-Plus perturbations \cite{liberoplus}, were not included in the analysis. $\pi_{0.5}$ and VLANeXt were identified as SOTA model choices, with publicly reported high task success rates on the LIBERO-Plus benchmark (83.9\% overall average for VLANeXt \cite{wu2026vlanextrecipesbuildingstrong} and 85.7\% overall average for $\pi_{0.5}$ \cite{zhang2026worldactionmodelsgeneralize}). An additional strength of VLANeXt, is that the authors made the underlying model architectural and design choices by trialling many combinations and choosing a recipe that optimised LIBERO-Plus task success rate. OpenVLA-OFT was also included, as the highest-performing model reported within the LIBERO-Plus paper\footnote{$\pi_{0.5}$ and VLANeXt were not included in the original LIBERO-Plus analysis. LIBERO-Plus (2025) predates VLANeXt (2026)} \cite{liberoplus, kim2025finetuningvisionlanguageactionmodelsoptimizing}.

Complete evaluation was carried out for $\pi_{0.5}$ and VLANeXt. We only evaluate OpenVLA-OFT for the spatial suite, as TSR degraded to 67.9\%, and results reported in the LIBERO-Plus paper indicate that the goal and long suites in particular would see further degradation \cite{liberoplus}. For a fair comparison, checkpoints were chosen where models had been fine-tuned on all the LIBERO suites \emph{jointly}, i.e. we avoided using checkpoints where suite-specific fine-tuning had been carried out.

\begin{table*}[!t]
\centering
\footnotesize
\renewcommand{\arraystretch}{0.85}
\setlength{\tabcolsep}{3.2pt}

\caption{
Cells report the median task-level percentage change in the mean \emph{successful} trajectory metric from baseline (original LIBERO) to the perturbation condition, with the number of tasks (out of 10) in the suite exhibiting a significant increase shown in parentheses. Cells are shaded when a significant increase is observed in the majority of tasks ($\geq 6$), with shading intensity indicating the size of the median percentage increase ($0$--$5\%$, $5$--$15\%$, $15$--$30\%$ and $>30\%$). Decreases and effects significant in fewer than 6 tasks are left unshaded. Dur. = Duration ($s$), C. Path = Cartesian path length ($m$), J. Path = joint path length ($rad$), C. Jerk = mean Cartesian jerk ($m/s^3$), J. Jerk = mean joint jerk ($rad/s^3$), Grip. = Total gripper movement ($m$), P95 C. Jerk = P95 Cartesian jerk ($m/s^3$). Models are jointly fine-tuned across all LIBERO suites.}
\label{tab:behaviour-main}

\begin{tabular}{@{}cll ccccccc !{\vrule width 0.8pt} cc@{}}
\toprule
& & & \multicolumn{7}{c}{\textbf{Median task-level change (1.d.p) (\%) (\#sig)}} & \multicolumn{2}{c}{\textbf{Success}} \\
\cmidrule(lr){4-10} \cmidrule(l){11-12}
& \textbf{Model} & \textbf{Pert.} & Dur. & C. Path & J. Path & C. Jerk & J. Jerk & Grip. & P95 C. Jerk & TSR & $n$ \\
\midrule
\multirow{21}{*}{\rotatebox[origin=c]{90}{\textbf{LIBERO-Spatial}}} & \multirow{7}{*}{\shortstack[l]{\textbf{$\pi_{0.5}$}\\ Original TSR 98.8\%\\ Mean Pert.* TSR 90.3\%}} 
& Back. &\cellcolor[HTML]{FDCC8A}\textbf{+10.3\%}(9)&+5.9\%(5)&+8.7\%(5)& -1.6\%(0) & -0.8\%(1) &\cellcolor[HTML]{FC8D59}\textbf{+26.4\%}(7) &+3.8\%(0)& 97.1\% & 501/516 \\
 &  
 & Cam. &\cellcolor[HTML]{FC8D59}\textbf{+22.5\%}(10)&\cellcolor[HTML]{FDCC8A}\textbf{+6.1\%}(7)&\cellcolor[HTML]{FDCC8A}\textbf{+11.1\%}(6)&\cellcolor[HTML]{FC8D59}\textbf{+24.1\%}(7)&\cellcolor[HTML]{FC8D59}\textbf{+18.8\%}(7)&\cellcolor[HTML]{D7301F}\textcolor{white}{\textbf{+37.8\%}(9)}& \cellcolor[HTML]{D7301F}\textcolor{white}{\textbf{+48.7\%}(7)}& 69.7\% & 524/752 \\
 &  
 & Lan. &\cellcolor[HTML]{FDCC8A}\textbf{+6.7\%}(7)&+1.9\%(4)&+2.3\%(3)& -3.1\%(1) & -1.3\%(3) &+7.9\%(3) & -3.8\%(0) & 90.8\% & 708/780 \\
 &  
 & Light &\cellcolor[HTML]{FDCC8A}\textbf{+5.4\%}(7)&\cellcolor[HTML]{FEF0D9}\textbf{+1.9\%}(6)&\cellcolor[HTML]{FDCC8A}\textbf{+5.6\%}(7)& -3.1\%(1) & -1.9\%(2) & 0.0\%(2) & -3.9\%(0) & 98.6\% & 576/584 \\

 &  
 & Obj. &+2.4\%(5)& -1.3\%(2) & -3.6\%(1) & -1.0\%(1) & -1.2\%(1) & -0.6\%(0) & -1.6\%(1) & 97.9\% & 377/385 \\
 &  
 & Rob. &\cellcolor[HTML]{FDCC8A}\textbf{+12.0\%}(10)&+2.5\%(3)&\cellcolor[HTML]{FDCC8A}\textbf{+5.4\%}(6)&+8.1\%(5)&\cellcolor[HTML]{FDCC8A}\textbf{+9.3\%}(6)&+18.9\%(3) &\cellcolor[HTML]{FC8D59}\textbf{+21.5\%}(7)& 84.9\% & 594/700 \\
 &  
 & Sen. &\cellcolor[HTML]{FDCC8A}\textbf{+13.6\%}(10)&+2.3\%(5)&+4.6\%(5)&+6.6\%(4)&+8.0\%(5)&+3.5\%(3) &+13.9\%(5)& 93.2\% & 654/702 \\

\cmidrule(lr){2-12}
 & \multirow{7}{*}{\shortstack[l]{\textbf{VLANeXt}\\ Original TSR 99.2\%\\ Mean Pert.* TSR 81.9\%}} 
 
 & Back. &\cellcolor[HTML]{FDCC8A}\textbf{+6.3\%}(7)&+1.2\%(3)&+2.0\%(4)& -0.4\%(0) & -0.6\%(0) &+3.0\%(3) &+2.0\%(1)& 95.2\% & 491/516 \\
 &  
 & Cam. &\cellcolor[HTML]{FEF0D9}\textbf{+4.2\%}(7)& -3.0\%(0) & -1.3\%(2) &+3.2\%(5)&+0.1\%(2)&+2.5\%(0) &\cellcolor[HTML]{FDCC8A}\textbf{+12.8\%}(7)& 68.0\% & 511/752 \\
 &  
 & Lan. &\cellcolor[HTML]{FEF0D9}\textbf{+4.7\%}(7)& -3.4\%(1) & -2.6\%(0) & -8.1\%(1) & -9.1\%(0) &+0.7\%(0) & -5.0\%(0) & 75.1\% & 586/780 \\
 &  
 & Light &+1.0\%(2)& 0.0\%(1) &+0.3\%(0)& -0.8\%(0) &+0.8\%(0)&+1.0\%(1) &+1.9\%(0)& 97.4\% & 569/584 \\
 &  
 & Obj. &+1.3\%(1)&+0.1\%(0)&+1.0\%(0)&+1.8\%(0)&+1.2\%(0)& -0.2\%(1) &+4.9\%(0)& 88.6\% & 341/385 \\
 &  
 & Rob. &+4.5\%(5)& -0.6\%(0) &+3.9\%(3)&+6.8\%(3)&+5.1\%(1)&+12.0\%(0)&\cellcolor[HTML]{FC8D59}\textbf{+15.5\%}(6)& 61.6\% & 431/700 \\
 &  
 & Sen. & -1.9\%(1) & -5.5\%(0) & -4.2\%(0) &+3.9\%(5)&+1.8\%(2)& -1.4\%(1) &\cellcolor[HTML]{FDCC8A}\textbf{+10.3\%}(7)& 87.5\% & 614/702 \\
\cmidrule(lr){2-12}

 & \multirow{7}{*}{\shortstack[l]{\textbf{OpenVLA-OFT}\\ Original TSR 97.8\%\\ Mean Pert.* TSR 67.9\%}} 

 & Back. &+3.4\%(5)& -0.9\%(1) & -2.7\%(1) &+4.3\%(4)&+2.0\%(4)&+1.6\%(1) &\cellcolor[HTML]{FC8D59}\textbf{+18.3\%}(6)& 81.4\% & 420/516 \\
 &  
 & Cam. &\cellcolor[HTML]{FC8D59}\textbf{+15.6\%}(6)& -1.8\%(3) & -1.8\%(2) &\cellcolor[HTML]{D7301F}\textcolor{white}{\textbf{+35.3\%}(10)}&\cellcolor[HTML]{FC8D59}\textbf{+25.3\%}(8)&\cellcolor[HTML]{D7301F}\textcolor{white}{\textbf{+37.7\%}(8)}&\cellcolor[HTML]{D7301F}\textcolor{white}{\textbf{+68.7\%}(10)}& 43.4\% & 326/752 \\
 &  
 & Lan. & 0.0\%(3) &+0.7\%(5)&+1.5\%(2)&+0.5\%(3)&+2.1\%(3)& -0.4\%(1) & -0.1\%(2) & 87.6\% & 683/780 \\
 &  
 & Light & -0.1\%(4) &+1.0\%(3)&+1.1\%(4)&+4.8\%(4)&+2.0\%(4)&+0.9\%(4)& +1.1\%(3) & 91.1\% & 532/584 \\
 &  
 & Obj. &+0.4\%(0)& 0.0\%(0) & -5.5\%(0) &+5.5\%(4)&+1.3\%(4)&+5.3\%(0) &+11.2\%(5)& 80.5\% & 310/385 \\
 &  
 & Rob. &+21.9\%(4)& -1.3\%(3) & -1.3\%(2) &\cellcolor[HTML]{FC8D59}\textbf{+28.7\%}(10)&\cellcolor[HTML]{FC8D59}\textbf{+22.3\%}(9)&\cellcolor[HTML]{D7301F}\textcolor{white}{\textbf{+79.5\%}(8)}&\cellcolor[HTML]{D7301F}\textcolor{white}{\textbf{+54.7\%}(10)}& 21.1\% & 148/700 \\
 &  
 & Sen. &+4.1\%(4)& -5.4\%(2) & -1.3\%(2) &\cellcolor[HTML]{FC8D59}\textbf{+16.6\%}(10)&\cellcolor[HTML]{FDCC8A}\textbf{+9.4\%}(7)&\cellcolor[HTML]{FC8D59}\textbf{+16.6\%}(7)&\cellcolor[HTML]{D7301F}\textcolor{white}{\textbf{+30.2\%}(10)}& 70.2\% & 493/702 \\
\midrule
\multirow{12}{*}{\rotatebox[origin=c]{90}{\textbf{LIBERO-Goal}}} & \multirow{7}{*}{\shortstack[l]{\textbf{$\pi_{0.5}$}\\ Original TSR 99.2\%\\ Mean Pert.* TSR 81.5\%}} 

& Back. &\cellcolor[HTML]{FEF0D9}\textbf{+3.2\%}(6)&+0.6\%(4)&+0.6\%(1)& -2.4\%(2) & -0.8\%(2) &+2.4\%(3) & -2.1\%(3) & 91.6\% & 515/562 \\
 &  
 & Cam. &\cellcolor[HTML]{FDCC8A}\textbf{+12.8\%}(8)&+2.5\%(5)& -2.2\%(3) &\cellcolor[HTML]{FDCC8A}\textbf{+9.4\%}(6)&+4.9\%(5)&\cellcolor[HTML]{D7301F}\textcolor{white}{\textbf{+38.4\%}(8)}&\cellcolor[HTML]{FC8D59}\textbf{+16.6\%}(6)& 73.4\% & 599/816 \\
 &  
 & Lan. &+2.8\%(5)& 0.0\%(4) & -2.9\%(3) & -3.8\%(1) & -1.9\%(1) &+3.9\%(2)&+2.7\%(0)& 70.6\% & 579/820 \\
 &  
 & Light &+2.2\%(5)& -0.5\%(3) & -4.2\%(2) & -0.1\%(0) &+1.0\%(3)&+1.0\%(2) & -2.3\%(1) & 93.7\% & 523/558 \\
 &  
 & Obj. &+5.2\%(4)&+2.7\%(1)&+2.0\%(3)&+0.8\%(1)&+2.0\%(1)&+1.7\%(1) &+4.7\%(0)& 72.2\% & 307/425 \\
 &  
 & Rob. &+11.2\%(5)&+4.8\%(4)&+1.1\%(4)&+2.8\%(3)&+4.4\%(4)&+14.8\%(4) &\cellcolor[HTML]{FC8D59}\textbf{+15.8\%}(6)& 78.9\% & 645/818 \\
 &  
 & Sen. &+7.5\%(5)&+1.3\%(3)& -1.2\%(2) &+4.5\%(2)&+2.2\%(2)&+6.7\%(3) &+9.2\%(0)& 90.0\% & 682/758 \\
\cmidrule(lr){2-12}
 & \multirow{7}{*}{\shortstack[l]{\textbf{VLANeXt}\\ Original TSR 97.8\%\\ Mean Pert.* TSR 73.6\%}} 
 
 & Back. &\cellcolor[HTML]{FDCC8A}\textbf{+7.9\%}(8)&+0.8\%(3)&+1.5\%(2)& -1.1\%(1) & -0.4\%(0) &+5.5\%(4) &+2.4\%(1)& 85.6\% & 481/562 \\
 &  
 & Cam. &\cellcolor[HTML]{FDCC8A}\textbf{+9.1\%}(8)& -2.9\%(1) &+3.2\%(3)& 0.0\%(2) & -1.7\%(3) &\cellcolor[HTML]{FDCC8A}\textbf{+9.3\%}(6) &+6.4\%(3)& 66.9\% & 546/816 \\
 &  
 & Lan. &\cellcolor[HTML]{FC8D59}\textbf{+17.5\%}(6)&+2.9\%(3)&+6.6\%(5)& -5.0\%(2) & -6.2\%(2) &+13.2\%(5) &+1.5\%(1)& 50.2\% & 412/820 \\
 &  
 & Light &+1.7\%(1)& -0.6\%(4) & -0.6\%(3) &+0.9\%(0)& -1.6\%(1) &+1.9\%(2) & +1.3\%(0)& 91.0\% & 508/558 \\
 &  
 & Obj. &+2.1\%(0)&+0.8\%(1)&+0.4\%(1)&+1.2\%(0)&+1.6\%(0)& -0.5\%(0)& -2.7\%(0) & 62.8\% & 267/425 \\
 &  
 & Rob. &+9.6\%(5)&+0.9\%(3)&+6.3\%(5)&+5.9\%(4)&+3.0\%(4)&+14.4\%(2) &+9.8\%(5)& 72.4\% & 592/818 \\
 &  
 & Sen. &\cellcolor[HTML]{FDCC8A}\textbf{+6.7\%}(8)& -2.5\%(1) &+0.4\%(0)&+3.2\%(2)& -0.7\%(2) &+16.0\%(5) &+6.4\%(3)& 86.5\% & 656/758 \\
\midrule
\multirow{12}{*}{\rotatebox[origin=c]{90}{\textbf{LIBERO-Long}}} & \multirow{7}{*}{\shortstack[l]{\textbf{$\pi_{0.5}$}\\ Original TSR 94.6\%\\ Mean Pert.* TSR 83.4\%}} 

& Back. & -0.3\%(2) & -0.3\%(2) & -3.0\%(2) &\cellcolor[HTML]{FDCC8A}\textbf{+8.2\%}(6)&\cellcolor[HTML]{FDCC8A}\textbf{+8.3\%}(8)&+1.7\%(2)&\cellcolor[HTML]{FDCC8A}\textbf{+11.6\%}(7)& 96.9\% & 560/578 \\
 &  
 & Cam. &+7.0\%(4)&+0.9\%(1)& -0.2\%(0) &\cellcolor[HTML]{FC8D59}\textbf{+18.2\%}(9)&\cellcolor[HTML]{FC8D59}\textbf{+16.2\%}(9)&\cellcolor[HTML]{FC8D59}\textbf{+20.9\%}(7)&\cellcolor[HTML]{D7301F}\textcolor{white}{\textbf{+30.7\%}(9)}& 49.2\% & 412/838 \\
 &  
 & Lan. & -1.7\%(0) & -0.4\%(2) & -2.3\%(1) &+1.5\%(3)& -0.4\%(2) & -2.1\%(1) &+1.7\%(1)& 95.4\% & 731/766 \\
 &  
 & Light & -1.4\%(2) & -0.7\%(2) & -3.5\%(1) & -0.8\%(1) & -2.3\%(2) &+1.0\%(4) & -2.1\%(1) & 93.6\% & 513/548 \\
 &  
 & Obj. &+0.1\%(0)& -0.1\%(0) & -0.9\%(3) &+1.3\%(2)&+1.2\%(1)& -0.5\%(2) &+2.6\%(1)& 91.7\% & 286/312 \\
 &  
 & Rob. &+3.2\%(1)& -0.1\%(1) &+1.7\%(0)&\cellcolor[HTML]{FDCC8A}\textbf{+14.4\%}(9)&\cellcolor[HTML]{FDCC8A}\textbf{+12.8\%}(9)&+17.6\%(4)&\cellcolor[HTML]{FC8D59}\textbf{+21.0\%}(10)& 80.8\% & 635/786 \\
 &  
 & Sen. &+1.8\%(2)& -1.0\%(0) & -2.3\%(1) &\cellcolor[HTML]{FC8D59}\textbf{+16.9\%}(10)&\cellcolor[HTML]{FDCC8A}\textbf{+13.2\%}(9)&+3.7\%(3)&\cellcolor[HTML]{FC8D59}\textbf{+24.1\%}(10)& 76.4\% & 686/898 \\
\cmidrule(lr){2-12}
 & \multirow{7}{*}{\shortstack[l]{\textbf{VLANeXt}\\ Original TSR 88.4\%\\ Mean Pert.* TSR 68.1\%}} 
 
 & Back. &+1.6\%(0)& -0.9\%(0) &+0.5\%(1)&+2.7\%(3)&+4.1\%(5)&+3.3\%(1)&+6.2\%(4)& 79.4\% & 459/578 \\
 &  
 & Cam. &\cellcolor[HTML]{FDCC8A}\textbf{+10.6\%}(6)& -1.4\%(2) & -0.4\%(2) &+4.3\%(2)&+3.7\%(4)&+8.5\%(4)&+9.3\%(4)& 47.0\% & 394/838 \\
 &  
 & Lan. &+6.4\%(5)& -0.3\%(0) &+0.1\%(1)&+4.4\%(5)&+1.8\%(2)&+10.6\%(5)&+5.0\%(4)& 58.6\% & 449/766 \\
 &  
 & Light & -0.2\%(0) & -0.6\%(0) & -0.1\%(0) &+0.5\%(2)&+1.0\%(2)&+2.0\%(2) &+3.7\%(2)& 72.3\% & 396/548 \\
 &  
 & Obj. &+2.2\%(3)&+0.8\%(1)&+0.4\%(2)&+2.0\%(0)&+2.0\%(1)& -0.1\%(1) &+2.3\%(0)& 84.0\% & 262/312 \\
 &  
 & Rob. &+3.2\%(1)& 0.0\%(0) &+2.6\%(1)&+3.5\%(4)&+4.1\%(5)&+3.7\%(0)&\cellcolor[HTML]{FDCC8A}\textbf{+7.7\%}(8)& 63.9\% & 502/786 \\
 &  
 & Sen. &+3.1\%(0)& -3.2\%(0) & -1.7\%(1) &+2.2\%(3)&+1.3\%(4)&+3.5\%(2)&+7.5\%(2)& 71.3\% & 640/898 \\
\midrule
\multirow{12}{*}{\rotatebox[origin=c]{90}{\textbf{LIBERO-Object}}} & \multirow{7}{*}{\shortstack[l]{\textbf{$\pi_{0.5}$}\\ Original TSR 99.4\%\\ Mean Pert.* TSR 90.5\%}} 
& Back. & -0.7\%(2) & -1.9\%(1) & -4.8\%(0) & -2.4\%(0) & -3.2\%(2) & -7.7\%(0) & -3.2\%(0) & 99.8\% & 495/496 \\
 &  
 & Cam. &+5.0\%(5)& -2.7\%(2) & -6.8\%(0) &\cellcolor[HTML]{FC8D59}\textbf{+19.9\%}(10)&\cellcolor[HTML]{FC8D59}\textbf{+17.2\%}(8)& -1.6\%(3)&\cellcolor[HTML]{D7301F}\textcolor{white}{\textbf{+38.1\%}(9)}& 84.1\% & 666/792 \\
 &  
 & Lan. & -1.0\%(0) & -2.3\%(0) & -4.4\%(0) & -1.8\%(0) & -2.7\%(0) & -8.3\%(0) & -3.7\%(0) & 89.7\% & 635/708 \\
 &  
 & Light &+1.3\%(2)& -1.2\%(0) & -4.6\%(1) & -5.3\%(0) & -3.9\%(1) & -8.0\%(1) & -2.3\%(0) & 99.0\% & 588/594 \\
 &  
 & Obj. & +0.0\%(0) & -1.1\%(0) & -2.8\%(2) &+2.8\%(2)&+1.4\%(2)& -1.7\%(0) &+8.6\%(0)& 90.3\% & 364/403 \\
 &  
 & Rob. &\cellcolor[HTML]{FDCC8A}\textbf{+10.5\%}(7)&+0.1\%(3)&+4.6\%(3)&+9.5\%(5)&\cellcolor[HTML]{FC8D59}\textbf{+16.0\%}(6)&+55.0\%(4) &\cellcolor[HTML]{FC8D59}\textbf{+22.5\%}(8)& 74.7\% & 595/796 \\
 &  
 & Sen. &+3.3\%(4)& -2.7\%(0) & -6.2\%(0) &\cellcolor[HTML]{FDCC8A}\textbf{+12.8\%}(8)&\cellcolor[HTML]{FDCC8A}\textbf{+10.4\%}(7)& -4.5\%(0) &\cellcolor[HTML]{FC8D59}\textbf{+20.2\%}(8)& 95.6\% & 807/844 \\
\cmidrule(lr){2-12}

 & \multirow{7}{*}{\shortstack[l]{\textbf{VLANeXt}\\ Original TSR 100.0\%\\ Mean Pert.* TSR 85.8\%}} 
 
 & Back. &\cellcolor[HTML]{FDCC8A}\textbf{+7.4\%}(9)&+0.5\%(4)&+1.0\%(1)& -2.8\%(0) & -0.2\%(0) & +18.3\%(5) &+1.3\%(0)& 94.4\% & 468/496 \\
 &  
 & Cam. &\cellcolor[HTML]{FDCC8A}\textbf{+10.4\%}(10)& -1.7\%(1) &+0.3\%(1)&+5.2\%(4)&\cellcolor[HTML]{FDCC8A}\textbf{+5.5\%}(6)&\cellcolor[HTML]{D7301F}\textcolor{white}{\textbf{+43.0\%}(6)} &\cellcolor[HTML]{FDCC8A}\textbf{+12.8\%}(6)& 74.0\% & 586/792 \\
 &  
 & Lan. &\cellcolor[HTML]{FDCC8A}\textbf{+10.7\%}(8)& +1.0\%(1) &-2.3\%(2)&-3.8\%(0)&-5.2\%(1)&+31.4\%(5)&+1.5\%(0)& 84.2\% & 596/708 \\
 &  
 & Light &\cellcolor[HTML]{FEF0D9}\textbf{+3.4\%}(7)&+1.2\%(5)& +1.2\%(3) & -1.4\%(0) & +0.5\%(0) &+0.1\%(3)&-3.4\%(0)& 99.0\% & 588/594 \\
 &  
 & Obj. &+5.3\%(5)&+1.8\%(0)&+2.4\%(1)&+3.8\%(0)&+2.7\%(0)&+36.5\%(0) &+7.8\%(1)& 84.9\% & 342/403 \\
 &  
 & Rob. &\cellcolor[HTML]{FDCC8A}\textbf{+15.0\%}(9)&+4.7\%(3)&\cellcolor[HTML]{FDCC8A}\textbf{+8.0\%}(6)&+6.7\%(5)&\cellcolor[HTML]{FDCC8A}\textbf{+10.5\%}(8)&+99.0\%(4) &+16.3\%(5)& 70.2\% & 559/796 \\
 &  
 & Sen. &\cellcolor[HTML]{FDCC8A}\textbf{+6.0\%}(10)& -1.9\%(0) & -0.8\%(0) & -0.9\%(0) & -0.2\%(1) &\cellcolor[HTML]{FC8D59}\textbf{+21.9\%}(8)&+2.3\%(0)& 94.2\% & 795/844 \\
\bottomrule
\end{tabular}
{\vspace{2pt}\footnotesize \textsuperscript{*}\,Mean perturbation robustness is reported as the macro-average of the seven perturbation condition TSRs, so that each perturbation condition is given equal weight, given the differences in sample sizes.}

\end{table*}

\section{RESULTS}

We structure our analysis around our research questions and provide the results in Table \ref{tab:behaviour-main} for each LIBERO suite.

\subsection{How does the typical behaviour of \emph{successful} trajectories change under perturbation? (RQ1)}

\textbf{We find that successful trajectories are not behaviourally invariant to perturbations.} Perturbations, in particular to camera, robot and sensor, can substantially alter successful trajectory behaviour. However, this sensitivity is model-dependent. Notable examples from the Spatial suite include a 24.1\% (7/10 tasks significant) and 35.3\% (10/10 tasks significant) median increase in mean Cartesian jerk under the \emph{camera} perturbation, for the $\pi_{0.5}$ and OpenVLA-OFT models respectively. These changes were also seen for other metrics, with these perturbations frequently affecting the duration, mean joint jerk, mean Cartesian jerk, and gripper movement (Table \ref{tab:behaviour-main}). Additionally, this pattern is observed across the suites, where camera, robot and sensor perturbations were typically responsible for producing the most significant behavioural changes in successful trajectories for the $\pi_{0.5}$ model. VLANeXt generally showed smaller shifts. The robot and camera perturbations appear to have the most significant effect on the behaviour metrics of $\pi_{0.5}$ and VLANeXt models. This finding aligns with those presented in the LIBERO-Plus paper, that models struggle with perturbations which require strong spatial geometry understanding and proprioception \cite{liberoplus}. While the TSR of $\pi_{0.5}$ in particular on the Spatial suite under the camera and robot perturbations (69.7\% and 84.9\%) are higher than the models evaluated in the LIBERO-Plus paper \cite{liberoplus}, these perturbations are still having a significant effect on the behaviour metrics.

\subsection{Are perturbation-induced behavioural changes reflected by changes in the TSR? (RQ2)}

\textbf{We find that TSR degradation and behavioural change are moderately correlated for a number of metrics, but that changes in task success do not fully capture changes in successful execution behaviour.} To understand whether reductions in task success are associated with altered behaviour in successful trajectories, we run a Spearman rank correlation \cite{schober2018correlation} analysis across all model, suite and perturbation conditions ((VLANeXt and $\pi_{0.5}$) models x 7 perturbations x 4 suites + (OpenVLA-OFT) model x 7 perturbations x 1 suites, n=63). For each behavioural metric, we correlate percentage degradation in TSR with the median task-level percentage change in that metric. Following conventions, we interpret absolute magnitudes of 0.00-0.10 as negligible, 0.10-0.39 as weak, 0.40-0.69 as moderate, 0.70-0.89 as strong and 0.9-1.0 as very strong \cite{schober2018correlation}. We find moderate positive correlations between TSR degradation and total gripper movement ($\rho$=0.56, p $<$ 0.00001, n=63), P95 Cartesian jerk ($\rho$=0.54, p $<$ 0.001, n=63), duration ($\rho$=0.52, p $<$ 0.0001, n=63), mean Cartesian jerk ($\rho$=0.43, p $<$ 0.001, n=63) and mean joint jerk ($\rho$ = 0.4, p $<$ 0.001, n=63). \footnote{Cartesian path length and joint path length showed negligible and weak non-significant correlations ($\rho$=-0.03, p = 0.80, n=63 and $\rho$=0.2, p = 0.11, n=63, respectively. These metrics also exhibited comparatively little perturbation-induced behaviour change overall, Table \ref{tab:behaviour-main}. 
}.This indicates that the perturbation conditions associated with larger reductions in task success also tended to produce larger behavioural changes in successful trajectories. 

A complementary question is whether similar task success implies similar behavioural robustness across models. We find this is not necessarily the case. Under the same perturbation there are a number of examples where models have similar TSR while showing substantially different changes in their successful trajectory behaviour. Under the \emph{camera} perturbation, the $\pi_{0.5}$ model often achieved similar TSR to VLANeXt, while exhibiting substantially more behaviour degradation. For example, on the Spatial suite, $\pi_{0.5}$ and VLANeXt had very comparable TSR, 69.7\% and 68.0\% respectively. However, successful $\pi_{0.5}$ trajectories showed substantially greater increases in duration (+22.5\%), mean Cartesian jerk (+24.1\%) and gripper movement (+37.8\%) than VLANeXt (+4.2\%, +3.2\%, and +2.5\%, respectively). These increases were also significant across more tasks for $\pi_{0.5}$: 10/10, 7/10 and 9/10 tasks for duration, mean Cartesian jerk and gripper movement compared with 7/10, 5/10 and 0/10 for VLANeXt. 

Similar discrepancies between TSR and behavioural robustness were observed across the other suites (Table \ref{tab:behaviour-main}). On the Long suite, $\pi_{0.5}$ had a higher TSR than VLANeXt under background, camera, robot and sensor perturbations, yet exhibited larger increases in mean Cartesian jerk, mean joint jerk and P95 Cartesian jerk in each case.

\subsection{Do perturbations alter the behavioural \emph{consistency} of successful trajectories? (RQ3)}

We additionally examine whether perturbations affect the \emph{consistency} of successful behaviour. While Table \ref{tab:behaviour-main} characterises changes in typical behaviour, it does not capture changes to the variability. We therefore calculate the median absolute deviation (MAD) for each behavioural metric under the baseline and perturbation conditions. An increase in MAD indicates greater variability between successful executions, while a decrease indicates a more tightly clustered successful behaviour. We observe an interesting pattern, where the perturbation conditions which produced the largest shifts in typical successful behaviour also frequently increased behavioural variability (Table \ref{tab:mad}). Notably, the camera perturbation led to an increase in MAD for every evaluated model and suite across duration, mean Cartesian jerk and gripper movement, indicating that successful execution became less consistent in addition to shifting away from baseline behaviour.

Interestingly, several perturbations led to reduced MAD when compared with the unperturbed baseline, particularly the background, language, lighting and object conditions, indicating a narrowing of the successful behaviour distribution. For example, under the lighting perturbation in LIBERO-Goal, gripper movement MAD decreased by 28\% for $\pi_{0.5}$ and 21\% for VLANeXt. As we are selecting only the successful trajectories, it could be the case that the perturbations are narrowing the set of behaviours that enable success. For example, for a given pick and place task, it may be that the policy can succeed in two ways, either via clean execution which generates a smooth, short trajectory, or by recovering after a misgrasp, which results in a longer and jerkier trajectory. Under perturbation, it could be that recovery becomes less reliable, and these trajectories now fail. This would mean that the remaining successful episodes are disproportionately clean/direct trajectories. Alternatively, the perturbation itself may be causing the policy to generate a narrower range of responses. Distinguishing between these could make for interesting future work. 

\newcolumntype{Y}{>{\centering\arraybackslash}X}

\newcommand{\metricsep}{%
    \color{gray!45}\vrule width 0.4pt
}

\newlength{\metricwidth}
\setlength{\metricwidth}{0.18\columnwidth}

\newcommand{\twomodel}[2]{%
    \begin{tabular*}{\metricwidth}{@{\extracolsep{\fill}}cc@{}}
     #1 & #2 
    \end{tabular*}%
}

\begin{table}[t]
\centering
\caption{Median percentage change in median absolute deviation (MAD)
relative to baseline for successful trajectories. Positive values indicate increased variability; negative values indicate reduced variability. Cell colour indicates the magnitude and direction of change using symmetric thresholds: $25$--$49\%$, $50$--$99\%$, and $\geq100\%$, with darker shading indicating larger changes. Values $<25\%$ are unshaded. VLN = VLANeXt.}

\label{tab:mad}

\scriptsize
\setlength{\tabcolsep}{1.2pt}
\renewcommand{\arraystretch}{0.95}

\begin{tabularx}{\columnwidth}{
    c l
    *{3}{Y}
    !{\metricsep}
    *{3}{Y}
    !{\metricsep}
    *{3}{Y}
}

\toprule
&
&
\multicolumn{3}{c}{Dur. (\%)}
&
\multicolumn{3}{c}{Cart. jerk (\%)}
&
\multicolumn{3}{c}{Grip. (\%)} \\

\cmidrule(lr){3-5}
\cmidrule(lr){6-8}
\cmidrule(lr){9-11}

&
Pert.
& $\pi_{0.5}$ & VLN & OFT
& $\pi_{0.5}$ & VLN & OFT
& $\pi_{0.5}$ & VLN & OFT \\

\midrule

\multirow{7}{*}{\rotatebox[origin=c]{90}{\textbf{LIBERO-Spatial}}}
& Back.
& \cellcolor[HTML]{EFAAAA} +50 & \cellcolor[HTML]{F8DEDE} +27 & \cellcolor[HTML]{DCEAF6} -33
& \cellcolor[HTML]{F8DEDE} +27 & -18 & -9
& \cellcolor[HTML]{EFAAAA} +80 & \cellcolor[HTML]{F8DEDE} +34 & -21 \\

&
Cam.
& \cellcolor[HTML]{D97878}\textcolor{white}{+142} & \cellcolor[HTML]{EFAAAA} +58 & \cellcolor[HTML]{D97878}\textcolor{white}{+317}
& \cellcolor[HTML]{D97878}\textcolor{white}{+160} & +22 & \cellcolor[HTML]{D97878} +122
& \cellcolor[HTML]{D97878}\textcolor{white}{+335} & \cellcolor[HTML]{F8DEDE} +38 & \cellcolor[HTML]{D97878}\textcolor{white}{+452} \\

&
Lan.
& 0 & \cellcolor[HTML]{EFAAAA} +50 & \cellcolor[HTML]{DCEAF6} -29
& +14 & -1 & -19
& -1 & -2 & -13 \\

&
Light
& +6 & 0 & \cellcolor[HTML]{BBC6D1} -50
& +12 & -5 & \cellcolor[HTML]{DCEAF6} -28
& \cellcolor[HTML]{F8DEDE} +34 & \cellcolor[HTML]{DCEAF6} -25 & \cellcolor[HTML]{DCEAF6} -33 \\

&
Obj.
& \cellcolor[HTML]{DCEAF6} -25 & 0 & -15
& +3 & +6 & +19
& +4 & +4 & -1 \\

&
Rob.
& \cellcolor[HTML]{F8DEDE} +45 & \cellcolor[HTML]{EFAAAA} +55 & \cellcolor[HTML]{D97878}\textcolor{white}{+263}
& \cellcolor[HTML]{EFAAAA} +72 & \cellcolor[HTML]{F8DEDE} +47 & \cellcolor[HTML]{F8DEDE} +43
& \cellcolor[HTML]{F8DEDE} +39 & +14 & \cellcolor[HTML]{D97878}\textcolor{white}{+797} \\

&
Sen.
& \cellcolor[HTML]{F8DEDE} +25 & 0 & \cellcolor[HTML]{EFAAAA} +69
& \cellcolor[HTML]{EFAAAA} +64 & +19 & \cellcolor[HTML]{F8DEDE} +46
& +22 & -2 & \cellcolor[HTML]{F8DEDE} +40 \\

\midrule

&
&
\multicolumn{3}{c!{\metricsep}}{\twomodel{$\pi_{0.5}$}{VLN}}
&
\multicolumn{3}{c!{\metricsep}}{\twomodel{$\pi_{0.5}$}{VLN}}
&
\multicolumn{3}{c}{\twomodel{$\pi_{0.5}$}{VLN}}
\\[-1pt]

\cmidrule{1-11}
\addlinespace[1pt]

\multirow{7}{*}{\rotatebox[origin=c]{90}{\textbf{LIBERO-Goal}}}
& Back.
& \multicolumn{3}{c!{\metricsep}}{\twomodel{-13}{\cellcolor[HTML]{F8DEDE} +31}}
& \multicolumn{3}{c!{\metricsep}}{\twomodel{+4}{+5}}
& \multicolumn{3}{c}{\twomodel{-11}{-13}} \\

&
Cam.
& \multicolumn{3}{c!{\metricsep}}{\twomodel{\cellcolor[HTML]{EFAAAA} +50}{\cellcolor[HTML]{EFAAAA} +61}}
& \multicolumn{3}{c!{\metricsep}}{\twomodel{+8}{+11}}
& \multicolumn{3}{c}{\twomodel{\cellcolor[HTML]{D97878}\textcolor{white}{+177}}{\cellcolor[HTML]{D97878}\textcolor{white}{+119}}} \\

&
Lan.
& \multicolumn{3}{c!{\metricsep}}{\twomodel{-23}{\cellcolor[HTML]{D97878}\textcolor{white}{+175}}}
& \multicolumn{3}{c!{\metricsep}}{\twomodel{+24}{+4}}
& \multicolumn{3}{c}{\twomodel{-17}{\cellcolor[HTML]{F8DEDE} +42}} \\

&
Light
& \multicolumn{3}{c!{\metricsep}}{\twomodel{-19}{0}}
& \multicolumn{3}{c!{\metricsep}}{\twomodel{-14}{-7}}
& \multicolumn{3}{c}{\twomodel{\cellcolor[HTML]{DCEAF6}-28}{-21}} \\

&
Obj.
& \multicolumn{3}{c!{\metricsep}}{\twomodel{\cellcolor[HTML]{DCEAF6} -36}{-7}}
& \multicolumn{3}{c!{\metricsep}}{\twomodel{-3}{-13}}
& \multicolumn{3}{c}{\twomodel{-22}{\cellcolor[HTML]{F8DEDE} +28}} \\

&
Rob.
& \multicolumn{3}{c!{\metricsep}}{\twomodel{\cellcolor[HTML]{EFAAAA} +54}{\cellcolor[HTML]{EFAAAA} +90}}
& \multicolumn{3}{c!{\metricsep}}{\twomodel{\cellcolor[HTML]{EFAAAA} +54}{+19}}
& \multicolumn{3}{c}{\twomodel{+24}{\cellcolor[HTML]{F8DEDE} +45}} \\

&
Sen.
& \multicolumn{3}{c!{\metricsep}}{\twomodel{\cellcolor[HTML]{F8DEDE} +31}{\cellcolor[HTML]{EFAAAA} +50}}
& \multicolumn{3}{c!{\metricsep}}{\twomodel{+23}{+16}}
& \multicolumn{3}{c}{\twomodel{0}{+17}} \\

\midrule

&
&
\multicolumn{3}{c!{\metricsep}}{\twomodel{$\pi_{0.5}$}{VLN}}
&
\multicolumn{3}{c!{\metricsep}}{\twomodel{$\pi_{0.5}$}{VLN}}
&
\multicolumn{3}{c}{\twomodel{$\pi_{0.5}$}{VLN}}
\\[-1pt]

\cmidrule{1-11}
\addlinespace[1pt]

\multirow{7}{*}{\rotatebox[origin=c]{90}{\textbf{LIBERO-Long}}}
& Back.
& \multicolumn{3}{c!{\metricsep}}{\twomodel{-9}{\cellcolor[HTML]{F8DEDE} +31}}
& \multicolumn{3}{c!{\metricsep}}{\twomodel{-8}{+2}}
& \multicolumn{3}{c}{\twomodel{+7}{\cellcolor[HTML]{EFAAAA} +64}} \\

&
Cam.
& \multicolumn{3}{c!{\metricsep}}{\twomodel{\cellcolor[HTML]{EFAAAA} +75}{\cellcolor[HTML]{D97878}\textcolor{white}{+126}}}
& \multicolumn{3}{c!{\metricsep}}{\twomodel{\cellcolor[HTML]{F8DEDE} +36}{+17}}
& \multicolumn{3}{c}{\twomodel{\cellcolor[HTML]{D97878}\textcolor{white}{+122}}{\cellcolor[HTML]{EFAAAA} +54}} \\

&
Lan.
& \multicolumn{3}{c!{\metricsep}}{\twomodel{-20}{\cellcolor[HTML]{EFAAAA} +55}}
& \multicolumn{3}{c!{\metricsep}}{\twomodel{-13}{+13}}
& \multicolumn{3}{c}{\twomodel{\cellcolor[HTML]{DCEAF6} -25}{\cellcolor[HTML]{F8DEDE} +33}} \\

&
Light
& \multicolumn{3}{c!{\metricsep}}{\twomodel{\cellcolor[HTML]{DCEAF6}-32}{\cellcolor[HTML]{F8DEDE} +31}}
& \multicolumn{3}{c!{\metricsep}}{\twomodel{-11}{-3}}
& \multicolumn{3}{c}{\twomodel{\cellcolor[HTML]{DCEAF6} -25}{+18}} \\

&
Obj.
& \multicolumn{3}{c!{\metricsep}}{\twomodel{-23}{-2}}
& \multicolumn{3}{c!{\metricsep}}{\twomodel{-19}{-3}}
& \multicolumn{3}{c}{\twomodel{\cellcolor[HTML]{DCEAF6} -35}{-13}} \\

&
Rob.
& \multicolumn{3}{c!{\metricsep}}{\twomodel{\cellcolor[HTML]{F8DEDE} +38}{\cellcolor[HTML]{F8DEDE} +37}}
& \multicolumn{3}{c!{\metricsep}}{\twomodel{\cellcolor[HTML]{F8DEDE} +32}{\cellcolor[HTML]{F8DEDE} +34}}
& \multicolumn{3}{c}{\twomodel{\cellcolor[HTML]{D97878}\textcolor{white}{+122}}{\cellcolor[HTML]{F8DEDE} +48}} \\

&
Sen.
& \multicolumn{3}{c!{\metricsep}}{\twomodel{+12}{\cellcolor[HTML]{F8DEDE} +41}}
& \multicolumn{3}{c!{\metricsep}}{\twomodel{\cellcolor[HTML]{F8DEDE} +44}{+9}}
& \multicolumn{3}{c}{\twomodel{\cellcolor[HTML]{F8DEDE} +31}{\cellcolor[HTML]{F8DEDE} +32}} \\

\midrule

&
&
\multicolumn{3}{c!{\metricsep}}{\twomodel{$\pi_{0.5}$}{VLN}}
&
\multicolumn{3}{c!{\metricsep}}{\twomodel{$\pi_{0.5}$}{VLN}}
&
\multicolumn{3}{c}{\twomodel{$\pi_{0.5}$}{VLN}}
\\[-1pt]

\cmidrule{1-11}
\addlinespace[1pt]

\multirow{7}{*}{\rotatebox[origin=c]{90}{\textbf{LIBERO-Object}}}
& Back.
& \multicolumn{3}{c!{\metricsep}}{\twomodel{\cellcolor[HTML]{DCEAF6} -30}{\cellcolor[HTML]{F8DEDE} +46}}
& \multicolumn{3}{c!{\metricsep}}{\twomodel{-18}{+15}}
& \multicolumn{3}{c}{\twomodel{+4}{+7}} \\

&
Cam.
& \multicolumn{3}{c!{\metricsep}}{\twomodel{\cellcolor[HTML]{EFAAAA} +53}{\cellcolor[HTML]{D97878}\textcolor{white}{+214}}}
& \multicolumn{3}{c!{\metricsep}}{\twomodel{\cellcolor[HTML]{EFAAAA} +59}{\cellcolor[HTML]{EFAAAA} +54}}
& \multicolumn{3}{c}{\twomodel{\cellcolor[HTML]{F8DEDE} +49}{\cellcolor[HTML]{D97878}\textcolor{white}{+266}}} \\

&
Lan.
& \multicolumn{3}{c!{\metricsep}}{\twomodel{\cellcolor[HTML]{DCEAF6} -41}{\cellcolor[HTML]{EFAAAA} +75}}
& \multicolumn{3}{c!{\metricsep}}{\twomodel{-8}{\cellcolor[HTML]{F8DEDE} +35}}
& \multicolumn{3}{c}{\twomodel{-2}{\cellcolor[HTML]{EFAAAA} +71}} \\

&
Light
& \multicolumn{3}{c!{\metricsep}}{\twomodel{-22}{-8}}
& \multicolumn{3}{c!{\metricsep}}{\twomodel{-10}{-2}}
& \multicolumn{3}{c}{\twomodel{+1}{-8}} \\

&
Obj.
& \multicolumn{3}{c!{\metricsep}}{\twomodel{+18}{\cellcolor[HTML]{F8DEDE} +42}}
& \multicolumn{3}{c!{\metricsep}}{\twomodel{+19}{\cellcolor[HTML]{F8DEDE} +40}}
& \multicolumn{3}{c}{\twomodel{+5}{\cellcolor[HTML]{F8DEDE} +47}} \\

&
Rob.
& \multicolumn{3}{c!{\metricsep}}{\twomodel{\cellcolor[HTML]{F8DEDE} +48}{\cellcolor[HTML]{D97878}\textcolor{white}{+225}}}
& \multicolumn{3}{c!{\metricsep}}{\twomodel{\cellcolor[HTML]{F8DEDE} +47}{\cellcolor[HTML]{EFAAAA} +84}}
& \multicolumn{3}{c}{\twomodel{\cellcolor[HTML]{D97878}\textcolor{white}{+125}}{\cellcolor[HTML]{D97878}\textcolor{white}{+258}}} \\

&
Sen.
& \multicolumn{3}{c!{\metricsep}}{\twomodel{+11}{\cellcolor[HTML]{EFAAAA} +67}}
& \multicolumn{3}{c!{\metricsep}}{\twomodel{\cellcolor[HTML]{EFAAAA} +67}{\cellcolor[HTML]{F8DEDE} +25}}
& \multicolumn{3}{c}{\twomodel{\cellcolor[HTML]{DCEAF6} -28}{+14}} \\

\bottomrule

\end{tabularx}
\end{table}

\section{LIMITATIONS AND FUTURE WORK}

The focus of this work is identifying whether \emph{successful} trajectories are meaningfully affected by input perturbations. In order to evaluate this, it was important to choose VLA models with high enough TSR that there were enough successful trajectories to evaluate, which is why full evaluation was only carried out for the $\pi_{0.5}$ and VLANeXt models. This focus also meant that investigating how perturbations affected failed trajectories was out of scope. Interesting questions for future work, include whether perturbations result in failed trajectories with meaningfully degraded behaviour.

The behavioural metrics we investigate are high-level and task-agnostic, allowing the methodology to be widely applied. However, this generality reduces the semantic specificity of the metrics. For example, increased gripper movement may act as a proxy for behaviours such as repeated grasping attempts or dropping and re-grasping objects. Diagnosing these events precisely would require task-and benchmark-specific adaptations to the evaluation.

Our experiments focus on simulated robotic manipulation tasks using LIBERO and LIBERO-Plus, reflecting the widespread use of these benchmarks for evaluating VLA models \cite{kim2024openvlaopensourcevisionlanguageactionmodel, black2026pi0visionlanguageactionflowmodel, kim2025finetuningvisionlanguageactionmodelsoptimizing}. This simulation enables controlled and repeatable comparison across many perturbation conditions and tasks. However, future work should establish whether the behavioural effects on successful trajectories observed here extend to physical robot deployment.

We have integrated our evaluation framework directly into VLA-eval~\cite{choi2026vlaeval}, an open-source and widely accessible evaluation framework for VLAs. In this way, we aim to provide the community with a toolkit that can be easily replicated and that can be applied to other popular evaluation benchmarks, in order to derive a more comprehensive ``model card" for VLA models~\cite{mitchell2019modelcard}.

\section{CONCLUSIONS}

In this work, we present a framework for evaluating the \emph{behavioural} robustness of VLA models, to complement task success rate. Our evaluation of SOTA VLA models on the LIBERO and LIBERO-Plus benchmarks demonstrates that perturbations can alter the behaviour of \emph{successful} trajectories, in a way which is not captured by task success rate alone. We encourage embodied AI benchmark developers to consider reporting metrics beyond task success rate alone, a stance motivated by our finding that doing so may be masking the impact perturbations have on behaviour observed within robotic manipulation trajectories. Therefore, we aim to offer the community a blueprint for model evaluation that can provide a more comprehensive report of model capabilities that reflects VLA models' robustness under input perturbations.







\bibliography{references}

\end{document}